%% file: main1.tex
\documentclass{article}

\usepackage{microtype}
\usepackage{graphicx}
\usepackage{subcaption}
\usepackage{booktabs} % for professional tables

\usepackage{hyperref}

\usepackage[dvipsnames]{xcolor}
\usepackage{color}
\usepackage{colortbl}
\usepackage{enumitem}
\usepackage{hyperref} 
\usepackage{multirow}
\usepackage{multicol}
\usepackage{graphicx}
\usepackage{caption}

\usepackage[most]{tcolorbox}
\newtcblisting{promptbox}[1]{
  breakable,
  enhanced,
  colback=black!2,
  colframe=black!20,
  boxrule=0.4pt,
  arc=2pt,
  left=6pt,
  right=6pt,
  top=6pt,
  bottom=6pt,
  title={#1},
  fonttitle=\bfseries,
  listing only,
  listing options={
    basicstyle=\ttfamily\footnotesize,
    breaklines=true,
    columns=fullflexible,
    keepspaces=true
  }
}

\usepackage[preprint]{icml2026}

\usepackage{amsmath}
\usepackage{amssymb}
\usepackage{mathtools}
\usepackage{amsthm}
\usepackage{enumerate}

\usepackage[capitalize,noabbrev]{cleveref}

\theoremstyle{plain}

\theoremstyle{definition}

\theoremstyle{remark}

\usepackage[textsize=tiny]{todonotes}

\icmltitlerunning{Graph via Vision for Structural Reasoning in LLMs}

\begin{document}

\twocolumn[
  \icmltitle{Visual Graph Scaffolds for Structural Reasoning in Large Language Models}



  \icmlsetsymbol{equal}{*}

    \begin{icmlauthorlist}
      \icmlauthor{Runlin Lei}{ruc}
      \icmlauthor{Xiaokui Xiao}{nus}
      \icmlauthor{Zhewei Wei}{ruc}
    \end{icmlauthorlist}
    
    \icmlaffiliation{ruc}{Renmin University of China}
    \icmlaffiliation{nus}{National University of Singapore}
    
    \icmlcorrespondingauthor{Zhewei Wei}{zhewei@ruc.edu.cn}
    
  \icmlkeywords{Machine Learning, ICML}

  \vskip 0.3in
]



\printAffiliationsAndNotice{}  % no special notice (required even if empty)

\begin{abstract}
\input{sections/abstract}
\end{abstract}

\input{sections/intro}
\input{sections/related}
\input{sections/method}
\input{sections/experiment}
\input{sections/limitation}
\input{sections/conclusion}

\bibliography{references}
\bibliographystyle{icml2026}

%%%%%%%%%%%%%%%%%%%%%%%%%%%%%%%%%%%%%%%%%%%%%%%%%%%%%%%%%%%%%%%%%%%%%%%%%%%%%%%
%%%%%%%%%%%%%%%%%%%%%%%%%%%%%%%%%%%%%%%%%%%%%%%%%%%%%%%%%%%%%%%%%%%%%%%%%%%%%%%
% APPENDIX
%%%%%%%%%%%%%%%%%%%%%%%%%%%%%%%%%%%%%%%%%%%%%%%%%%%%%%%%%%%%%%%%%%%%%%%%%%%%%%%
%%%%%%%%%%%%%%%%%%%%%%%%%%%%%%%%%%%%%%%%%%%%%%%%%%%%%%%%%%%%%%%%%%%%%%%%%%%%%%%
\newpage
\appendix
\onecolumn
\input{sections/appendix}

\end{document}

%% file: sections/abstract.tex
Graphs have been used to enhance large language models (LLMs) for structured reasoning, mostly as external knowledge sources are provided to models at test time. 
In this paper, we take a different view: the value of graphs for LLMs lie not only in supplying information, but also in organizing reasoning. 
Inspired by how humans use graph-structured mind maps to organize branching and converging thoughts, we ask whether graphs can serve as an internal form of reasoning assistance. 
We study this question on multi-hop question answering tasks, where teacher-provided reasoning traces are rewritten as graph mind maps and used to guide a student model. 
Our experiments reveal a clear modality gap. 
When graph structures are flattened into text, their benefits become limited once direct answer hints are removed. 
Under this abstract guidance setting, both reasoning efficiency and answer quality degrade substantially. 
In contrast, visual graph guidance remains effective without direct answer clues, and its advantage persists after supervised fine-tuning and KL-based distillation. 
The above findings support the claim that graphs should be studied not only as external knowledge structures for LLMs, but also as visual scaffolds for organizing reasoning.

%% file: sections/intro.tex
\section{Introduction}

Graphs can serve as useful tools for enhancing Large Language Models (LLMs and Vision Language Models (VLMs) in reasoning tasks. 
In most existing settings, graphs are used as external support, where they retrieve evidence, ground answers, or organize memories that the model may not possess~\cite{han2024retrieval, g-retrieve, zhang2025g-memory}. 
While effective, this view captures only part of what graphs can offer. 
In human reasoning, graphs often function not only as information structures, but also as cognitive scaffolds. 
For example, human write mind maps that make branching, convergence, hierarchy, and local relations easier to inspect than linear text. 
This motivates the central question of this paper: \textit{can graphs help LLMs not only access knowledge, but also organize reasoning?}

We study this question in a teacher--student setting. 
A stronger teacher model first solves a multi-hop question answering problem, and its reasoning process is rewritten into a graph-structured scaffold for a weaker student model. 
The goal is not to retrieve additional facts, but to transfer the organization of a successful reasoning process. 
If such guidance can improve the student and later be internalized through fine-tuning or distillation, then graphs may serve as not only external knowledge.
Instead, they become a medium for teaching structured thought.

A natural way to implement graph-structured reasoning is through text. 
Prior work such as Graph-of-Thoughts and related methods has explored how non-linear reasoning structures can be represented within language-based prompting frameworks~\cite{got, got2, han_reaosning_with_graphs}. 
However, text remains a linear medium. 
Once a graph is flattened into sentences, its topology must be described indirectly, often making the guidance more redundant, and harder to learn from. 
This motivates our alternative interface: \textbf{graph as image}. 
In our pipeline, teacher reasoning is rendered as a graph-structured mind map and provided to a student VLM, while textual guidance serve as controlled baselines. 
This design lets us ask whether the benefit comes from reasoning content alone, or from preserving reasoning topology in a visual form.

\begin{figure*}[ht]
    \centering
    \includegraphics[width=1\linewidth]{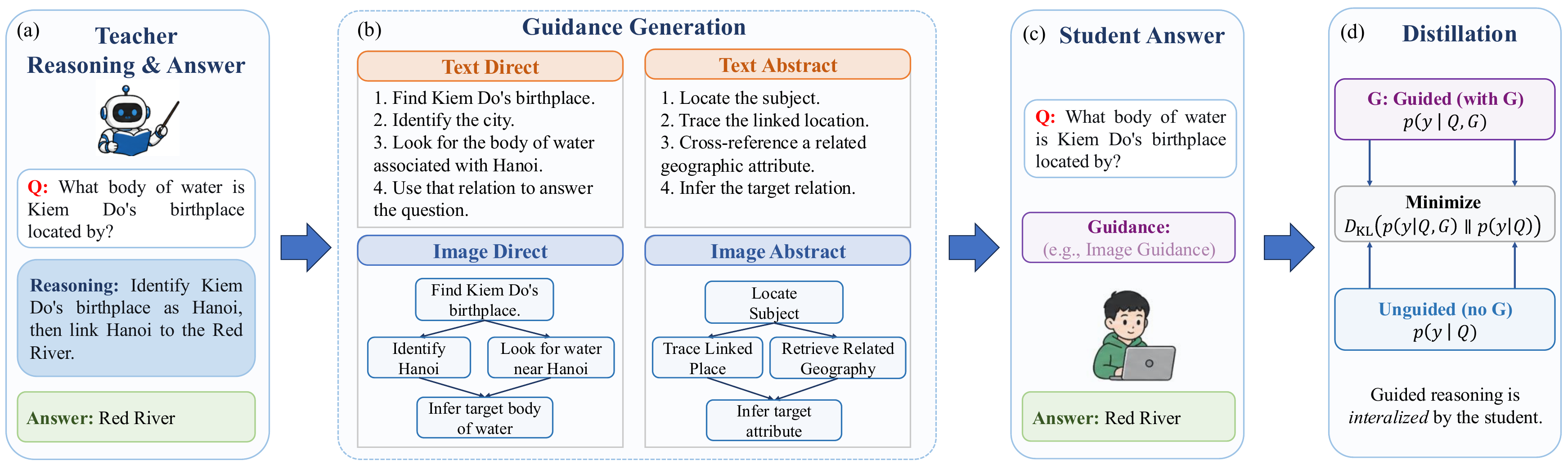}
    \caption{Overview of the Graph-Guided Reasoning framework. \textbf{(a)} Teacher Reasoning \& Answer: A strong teacher model solves a multi-hop question and generates a detailed reasoning trace. \textbf{(b)} Guidance Generation: The teacher's reasoning is transformed into four types of guidance artifacts: Textual vs. Visual modalities, and Direct vs. Abstract styles. \textbf{(c)} Student Answer: A student model utilizes the generated guidance to arrive at the correct answer. \textbf{(d)} Distillation: The student internalizes the structured reasoning via SFT or KL.}
    \label{fig:full}
    \vspace{-0.5em}
\end{figure*}

To make this comparison meaningful, we distinguish between two guidance settings. 
In the direct setting, guidance may contain answer-local hints such as key facts or intermediate conclusions. 
In the abstract setting, such hints are forbidden: the guidance may only describe general reasoning strategies and structural relations, without leaking the final answer, answer-specific facts, or intermediate conclusions. 
The abstract setting is central to our study because it tests whether the student can use the graph as a reasoning scaffold rather than as a shortcut to the answer.
Our experiments reveal a clear modality gap. 
When answer-local hints are allowed, visual graph guidance and textual guidance perform similarly. 
When guidance must remain abstract, however, visual graph guidance remains effective while textual guidance degrades sharply. 
This advantage also persists after supervised fine-tuning (SFT) and KL-based distillation, and is accompanied by shorter reasoning outputs. 

These findings suggest a broader role for graphs in graph--LLM. 
Beyond serving as external knowledge, graphs can act as interfaces for transferring the organization of reasoning itself. 
By preserving branching, convergence, and local dependencies in a compact form, visual graphs expose structure that is difficult to maintain when the same reasoning process is flattened into text. 
Our position is that visual graph guidance should be treated as a topology-preserving interface for structured reasoning, not merely as another form of prompting. 
In this interface, vision is a particularly promising modality because it can preserve and present graph topology instead of serializing it linearly into text. 

%% file: sections/related.tex
\section{Related Work}

Many graph-LLM works treat graphs as external structures for retrieval or grounding. 
GraphRAG constructs entity--relation graphs from corpora for retrieval over local facts and corpus-level structure~\citep{edge2024local}. 
G-Retriever retrieves compact, reasoning-relevant subgraphs before generation~\citep{g-retrieve}. 
ToG-2 alternates knowledge-graph traversal with textual context retrieval for deeper multi-hop reasoning~\citep{tog}, while GNN-RAG uses graph neural retrieval to identify question-relevant nodes and paths before prompting the LLM~\citep{gnnrag}. 
A related line uses graphs to organize reasoning itself: Graph of Thoughts represents intermediate reasoning states as nodes and dependencies as edges, enabling non-linear reasoning trajectories~\cite{got, got2}. 
Recent work on visual graph reasoning further shows that vision-language models can benefit from graph images.
Wang et al.~\cite{wang2023can} show that LLMs struggle with text-based graph reasoning, Zhao et al.~\cite{zhao2025underappreciated} find that vision encoders can outperform GNNs on global structural understanding, and Wei et al.~\cite{wei2024gita} and Zhu et al.~\cite{zhu2025benchmarking} use image-based representations of raw graph inputs to improve graph reasoning performance. 
However, these studies mostly use graphs as external knowledge, textual reasoning structures, or visualization of graph inputs. 
They leave underexplored whether a graph image can externalize and transfer the structure of a reasoning process itself. 
This paper focuses on that interface question: instead of asking whether a model can solve a graph problem from an image, we ask whether an image can carry a reasoning trajectory from one model to another. 

%% file: sections/method.tex
\section{Visual Graphs Are Reasoning Interfaces}

To study the differences caused by the presentation of graphs during reasoning in LLMs, we design a pipeline that compares different representations of the same teacher guidance, which specifically focus on the comparison between rendered graph images versus linear text.
As illustrated in Figure~\ref{fig:full}, the pipeline has three main stages: teacher trajectory generation, guidance construction, with distillation as an internalization step.
This design isolates whether the same reasoning structure is more usable when exposed visually rather than flattened into text.

\subsection{Teacher Trajectory Generation}

We first identify QA examples that the base student answers incorrectly. 
For each such case, a stronger teacher model solves the same question and produces an explicit reasoning trajectory. 
We keep only cases where the teacher's answer is verified as correct, so that the following comparison focuses on how the reasoning is transferred rather than whether the teacher solved the problem. 
These verified teacher trajectories are then rewritten into guidance artifacts for the student.

\subsection{Guidance Construction}

Each teacher trajectory is converted into guidance along two axes: modality and content style. 
For modality, \textbf{image} guidance converts the teacher trajectory into Graphviz DOT code and renders it as a graph-structured mind map, while \textbf{text} guidance expresses the same kind of support in plain text. 
We also construct a \textbf{graph-to-text} control, which converts the teacher-generated graph code into text after graph construction. 
This control preserves graph node content while removing visual layout.

For content style, \textbf{direct} guidance includes task-specific hints, key facts, and intermediate conclusions. 
In contrast, \textbf{abstract} guidance may include only general reasoning strategies and logical operations, and must exclude answer-specific clues.
Direct guidance is included because it provides answer-local information, while abstract guidance tests the student's ability to truly use the graph structure like a mind map to reason.

\subsection{Student Use of Guidance}

The student uses the constructed guidance in two ways. 
First, in \textbf{guided re-evaluation}, the student is frozen and re-answers its original failure cases with teacher guidance. 
This measures the immediate usefulness of each guidance interface. 
Second, in \textbf{internalization}, the student is trained on successful guided behavior and later tested without guidance. 
We consider both \textbf{Self-SFT}, which fine-tunes the student on its own correct guided responses, and \textbf{KL distillation}, where a guided branch provides soft targets to an unguided branch. 
Together, these settings ask whether visual graphs help only as inference-time prompts, or whether their structured signal can be absorbed into the model.

%% file: sections/experiment.tex
\section{Experiments}

We aim to study the below research questions:
\begin{itemize}[leftmargin=*]
    \item \textbf{RQ1:} Can visual graphs effectively guide the reasoning of student models?
    \item \textbf{RQ2:} Can student models internalize the advantages of graph-based guidance?
    \item \textbf{RQ3:} What role does graph structure play in visual reasoning guidance?
\end{itemize}
To answer these questions, we first perform guided re-evaluation on teacher-correct failure cases with a frozen student model.
We then examine whether the observed advantage persists after internalization through Self-SFT and KL distillation.
Finally, we analyze why images help by measuring output length and ablating graph topology.

\subsection{Experiment Setting}
The main experiments are conducted on three classic multi-hop QA datasets: HotpotQA~\cite{yang2018hotpotqa}, 2WikiMultiHopQA~\cite{2wiki}, and MuSiQue~\cite{trivedi2022musique}.
The supervision dataset is built from the training splits of these three datasets.
After semantic validation, it contains $14{,}490$ teacher-correct cases for guidance-based re-evaluation and downstream internalization.
The held-out QA test set contains $3{,}000$ questions in total, with $1{,}000$ sampled from each dataset.
For the ablation study, we use a separate $3{,}000$-example subset sampled from the training-split teacher-correct pool, again with $1{,}000$ examples per dataset.
We report two QA evaluations.
The first is guided re-evaluation on the teacher-guided failure set.
The second is QA on the test set after internalization.
In the experiments, we instantiate the teacher, student, and verifier with DeepSeek-V3.2~\cite{liu2025deepseek}, Qwen3-VL-8B-Instruct~\cite{bai2025qwen3}, and Qwen3-8B-Instruct~\cite{yang2025qwen3}, respectively.
Full prompts, setup details, and hyperparameters are deferred to the appendix.

\subsection{Main Results}
\begin{table}[h]
\centering
\caption{Main QA results.
Re-eval reports guided re-evaluation accuracy (\%) on teacher-correct QA failures with a frozen student.
Self-SFT and KL report held-out QA accuracy (\%) on the test set after internalization.
Train with teacher CoT stands for directly SFT on teacher's reasoning content.}
\label{tab:main_results}
\resizebox{\columnwidth}{!}{
\begin{tabular}{llcc}
\toprule
\textbf{Experiment} & \textbf{Guidance} & \textbf{Direct} & \textbf{Abstract} \\
\midrule
\multicolumn{4}{l}{\textit{Guided re-evaluation on the failure pool}} \\
\midrule
\multirow{4}{*}{Re-eval}
& Image guide & 71.21 & \textbf{70.80} \\
& Text guide & \textbf{71.22} & 51.97 \\
& Graph-to-text control & 70.74 & 46.40 \\
\midrule
\multicolumn{4}{l}{\textit{After internalization}} \\
\midrule
\multicolumn{2}{l}{Baseline with CoT} & \multicolumn{2}{c}{59.37} \\
\multicolumn{2}{l}{Train with teacher CoT} & \multicolumn{2}{c}{67.17} \\
\midrule
\multirow{3}{*}{Self-SFT}
& Image guide & \textbf{64.00} & 63.27 \\
& Text guide & 61.53 & 58.23 \\
& Graph-to-text control & 61.77 & 59.03 \\
\midrule
\multirow{2}{*}{KL distillation}
& Image guide & \textbf{64.40} & \textbf{64.47} \\
& Text guide & 63.40 & 58.37 \\
\bottomrule
\end{tabular}
}
\end{table}

\paragraph{Finding 1: \textbf{The image advantage appears when reasoning guidance must remain structural rather than answer-local.}}
Table~\ref{tab:main_results} shows that modality matters little in the direct setting.
During the re-evaluation, direct image and direct text guidance are essentially tied.
This is consistent with the idea that once strong answer-local hints are present, changing the modality matters little.
The pattern changes in the abstract setting.
Here the guidance must teach how to reason rather than reveal what answer to recover.
Under this constraint, abstract image guidance remains strong while abstract text falls largely with the Graph-to-text control drops even further.
These results suggest that images become especially valuable when the student must rely on structural guidance rather than answer-local clues.

\paragraph{Finding 2: \textbf{This advantage survives internalization.}}
The Self-SFT and KL blocks in Table~\ref{tab:main_results} show that the same ordering persists after training.
In Self-SFT, image guidance remains stronger than text in both styles.
KL distillation shows the same pattern, with the largest gap again appearing in the abstract setting.
The graph-to-text control again remains below image guidance.
This indicates that image-based graph-structured guidance is easier for the student to internalize than its text counterpart.

\subsection{Why Images Help}
\paragraph{Finding 3: \textbf{Images help by offering a shorter but more structured interface.}}
Table~\ref{tab:tokens} reports average output length in the abstract setting.
In re-evaluation, abstract image guidance yields only $226$ output tokens on average, compared with $703$ for abstract text and $697$ for graph-to-text.
After training, the abstract image models also remain far shorter than their text counterparts.
This suggests that the graph representation provides a strong compression effect.
Rather than spending tokens on unpacking a long sequential text guidance, the model can recover the reasoning process from a more concise structural interface via vision.
In this sense, the rendered image preserves the compactness with which graphs express complex relations while foregrounding the reasoning pattern.

\begin{table}[ht]
\centering
\caption{Average output tokens in the abstract setting.
Image guidance stays substantially shorter than text-based alternatives.}
\label{tab:tokens}
\begin{tabular}{lccc}
\toprule
\textbf{Setting} & \textbf{Image} & \textbf{Text} & \textbf{Graph-to-text} \\
\midrule
Re-eval  & 226 & 703 & 697 \\
Self-SFT  & 377 & 780 & 810 \\
KL & 360 & 844 & -- \\
\bottomrule
\end{tabular}
\end{table}

\paragraph{Finding 4: \textbf{Topology preservation is critical for abstract visual guidance.}}
In Table~\ref{tab:ablation}, we conduct an ablation study to analyze the effect of structure during image-guidance.
We can see that when abstract image guidance is forced into a chain, or when its node budget is sharply reduced before rendering, re-evaluation accuracy drops substantially.
The drop points to the role of preserved branching and convergence in abstract guidance, rather than to visual presentation alone. 
In this setting, graph topology appears to carry part of the supervision signal.

\begin{table}[ht]
\centering
\caption{Abstract image-guidance ablations on guided re-evaluation accuracy (\%).}
\label{tab:ablation}
\begin{tabular}{lcc}
\toprule
\textbf{Variant} & \textbf{Accuracy} & \textbf{$\Delta$} \\
\midrule
Baseline & 72.23 & -- \\
\midrule
Chain & 57.82 & -14.41 \\
5 nodes & 55.96 & -16.27 \\
10 nodes & 56.09 & -16.14 \\
\bottomrule
\end{tabular}
\vspace{-0.5em}
\end{table}

%% file: sections/limitation.tex
\section{Limitations}

\paragraph{Limitation 1: \textbf{Transfer beyond the QA family remains limited.}}
The image advantage is clearest within the QA family on which guidance is constructed and the student is trained. 
Although the same ordering holds on a separate six-dataset reasoning benchmark, the absolute performance remains modest as shown  in Table~\ref{tab:limitation_reasoning}. 
This suggests that the current pipeline is still task-specific and does not yet establish a broadly reusable structured-reasoning capability.

\begin{table}[h]
\centering
\caption{Out-of-domain reasoning as a limitation signal.}
\label{tab:limitation_reasoning}
\begin{tabular}{lc}
\toprule
\textbf{Model} & \textbf{Accuracy} \\
\midrule
Baseline + CoT & 47.33 \\
\midrule
Image guidance & 39.15 \\
Text guidance & 28.42 \\
\bottomrule
\end{tabular}
\end{table}

A stronger test would require guidance construction and training across more diverse reasoning families, rather than transfer from multi-hop QA alone.

\paragraph{Limitation 2: \textbf{Image supervision does not yet match direct CoT supervision.}}
As Table~\ref{tab:main_results} shows, image guidance is stronger than text guidance within the constrained guidance pipeline studied here, but it still does not outperform direct training on teacher's CoT.
The contribution should therefore be read as a claim about a better interface for transferring structure under restricted supervision, not as evidence that image guidance is a complete replacement for current distillation for reasoning.

%% file: sections/conclusion.tex
\section{Conclusion}
This paper argues that graphs should be studied not only as external knowledge for LLMs, but also as topology-preserving interfaces for organizing reasoning. 
We find that visual graph guidance matches text guidance when answer-local hints are available, but remains far more effective when guidance must stay abstract. 
This advantage persists after internalization and weakens when graph is disrupted, suggesting that the useful signal lies in preserved structure rather than visual presentation alone.
A broader take-away is that graph--LLM integration should not be framed only as retrieval or grounding. 
Graphs can serve as compact abstractions for reasoning organization, and vision offers a promising modality for exposing this structure without flattening it into text. 

%% file: sections/appendix.tex
\section{Additional Details}

\subsection{Experimental Setup}

\paragraph{Classic QA supervision pool.}
The main pipeline is built on the training splits of HotpotQA ($20{,}000$ examples), 2WikiMultiHopQA ($20{,}000$), and MuSiQue ($19{,}938$).
The base student is first evaluated on these training questions with a direct-answer prompt.
Teacher generation is then applied to the student-failure pool.
After semantic validation, this yields $18{,}785$ usable teacher records, of which $14{,}490$ are teacher-correct.
These teacher-correct cases are the source of guided re-evaluation and downstream internalization.
The number of evaluable examples can differ across guidance conditions because image guidance requires valid graph code and successful rendering, whereas text guidance only requires non-empty textual guidance.
Consequently, the re-evaluation accuracy in Table~\ref{tab:main_results} is computed with the condition-specific evaluable denominator, not always with $14{,}490$ as the denominator.

\paragraph{Held-out QA test set.}
The held-out QA test set contains $3{,}000$ examples in total.
It is constructed by sampling $1{,}000$ test questions from each of HotpotQA~\cite{yang2018hotpotqa}, 2WikiMultiHopQA~\cite{2wiki}, and MuSiQue~\cite{trivedi2022musique}.
All main QA results in the paper use this fixed test set.

\paragraph{Ablation diagnostic subset.}
The ablation study uses a separate $3{,}000$-example subset sampled from the training-split teacher-correct pool, with $1{,}000$ examples from each QA dataset.
This subset is not the held-out QA test set.
It is used because the ablation experiment must regenerate teacher graph code under alternative structural constraints and then re-evaluate the frozen student on the same kind of teacher-guided failure cases as the main re-evaluation experiment.
After filtering to examples with valid guidance, the abstract-image ablation baseline contains $2{,}827$ examples.

\paragraph{Reasoning benchmark construction.}
The out-of-domain reasoning benchmark is constructed separately from six reasoning datasets:
CLUTRR~\cite{sinha2019clutrr},
LogiQA~\cite{liu2020logiqa},
AR-LSAT~\cite{zhong2021arlsat},
AIW-Easy~\cite{nezhurina2024alice},
AIW-Hard~\cite{nezhurina2024alice},
and Bamboogle~\cite{bamboo}.
Unlike the QA test set, this benchmark is not re-sampled to a fixed per-dataset size.
The resulting benchmark contains $2{,}544$ examples in total.
Its dataset composition is shown in Table~\ref{tab:reasoning_breakdown}.

\begin{table}[h]
\centering
\caption{Construction of the out-of-domain reasoning benchmark.}
\label{tab:reasoning_breakdown}
\begin{tabular}{lr}
\toprule
\textbf{Dataset} & \textbf{Examples} \\
\midrule
CLUTRR & 1146 \\
LogiQA & 651 \\
AR-LSAT & 230 \\
AIW-Easy & 200 \\
AIW-Hard & 192 \\
Bamboogle & 125 \\
\midrule
Total & 2544 \\
\bottomrule
\end{tabular}
\end{table}

Three details matter for interpreting this benchmark.
First, the reasoning benchmark is used only for evaluation, not for the main guidance-construction claim.
Second, CLUTRR, LogiQA, and AR-LSAT use the context-present prompt branch, whereas AIW-Easy, AIW-Hard, and Bamboogle are treated as no-context datasets.
Third, LogiQA and AR-LSAT additionally require the final answer to include both the option letter and the full option text.

\paragraph{Prompting, rendering, and decoding.}
The base QA failure pool is created with a standard non-chain-of-thought QA prompt and no external guidance.
Guided re-evaluation, held-out QA testing, and the downstream training data used in Self-SFT and KL all use chain-of-thought QA prompts.
For image guidance, the teacher first generates Graphviz DOT graph code.
That graph code is then rendered at $8\times 8$ inches and 150 DPI, and the resulting image is resized to $1024\times1024$.
Inference uses temperature $0.2$, top-$p=0.95$.
The teacher-generation API is called with temperature $0.1$.
The verifier uses Qwen3-8B~\cite{yang2025qwen3}.

\paragraph{Training setup.}
Train with teacher CoT uses all $14{,}490$ teacher-correct QA cases.
It always targets the teacher's original round-1 chain-of-thought response, even when correctness was confirmed by the later semantic-validation step.
The correct re-evaluation responses used for downstream Self-SFT and KL training are condition-specific.
Table~\ref{tab:reeval_denominators} reports the corresponding denominators and correct-response counts.

\begin{table}[h]
\centering
\caption{Condition-specific guided re-evaluation denominators and training-response counts.
The number of correct guided responses is also the number of responses used by the corresponding Self-SFT/KL condition.}
\label{tab:reeval_denominators}
\begin{tabular}{lrrr}
\toprule
\textbf{Guidance condition} & \textbf{Evaluable} & \textbf{Correct} & \textbf{Accuracy} \\
\midrule
Image direct & 14,452 & 10,291 & 71.21 \\
Image abstract & 13,388 & 9,479 & 70.80 \\
Text direct & 14,467 & 10,303 & 71.22 \\
Text abstract & 14,490 & 7,530 & 51.97 \\
Graph-to-text direct & 14,452 & 10,224 & 70.74 \\
Graph-to-text abstract & 14,484 & 6,720 & 46.40 \\
\bottomrule
\end{tabular}
\end{table}

\paragraph{Optimization details.}
Both Self-SFT and KL use LoRA rank $32$, LoRA alpha $64$, dropout $0.05$, learning rate $10^{-5}$, weight decay $0.01$, warmup ratio $0.05$, batch size $2$, gradient accumulation $32$, and maximum sequence length $4096$.
The LoRA adapters are attached to \texttt{q\_proj}, \texttt{k\_proj}, \texttt{v\_proj}, \texttt{o\_proj}, \texttt{gate\_proj}, \texttt{up\_proj}, and \texttt{down\_proj}.
KL distillation uses temperature $\tau=2.0$ and the three-epoch schedule $\alpha=\{1.0,0.5,0.0\}$.

\subsection{Prompt Templates}
This section documents the prompt templates used in the experiments.
We show the templates with placeholders such as \texttt{\textless question\textgreater}, \texttt{\textless context\textgreater}, and \texttt{\textless text\_guidance\textgreater}.
Unless noted otherwise, the boxes below show the context-present branch used by the three QA datasets in the main experiments.
When a dataset has no usable context, the corresponding \texttt{Context:} block is omitted.
For multiple-choice datasets such as LogiQA and AR-LSAT, we append one extra instruction requiring both the option letter and the full option text in the final answer.

\subsubsection{Base QA and teacher verification}
The base QA failure pool is created with a direct answer prompt rather than a chain-of-thought prompt.
Teacher supervision is then built with a chain-of-thought teacher prompt and, when needed, a semantic validation turn.

\begin{promptbox}{Base student QA prompt used to create the failure pool}
Context:
<context>

Question: <question>

Please answer the question based on the given context,
and provide your final answer in this exact format:
Answer: <your answer>.

Make sure to end with "Answer:" followed by your final answer.
\end{promptbox}

\begin{promptbox}{Teacher Round-1 CoT QA prompt}
Context:
<context>

Question: <question>

Think through this problem step by step before giving your final answer.

Format your response as:
Reasoning: <your step-by-step analysis>
Answer: <your final answer>
\end{promptbox}

\begin{promptbox}{Teacher semantic validation prompt}
The ground truth answer is: <ground_truth>

Your answer above may differ from the ground truth in wording.

Task: Determine if YOUR answer is actually correct
(semantically equivalent to ground truth).

Consider:
1. Are they the same entity or concept with different names?
2. Are they semantically equivalent answers to the question?
3. Is your answer a valid response even if phrased differently?

Response format:
- Briefly explain why your answer is or is not correct
- End with exactly "Judgment: CORRECT"
  or "Judgment: INCORRECT"
\end{promptbox}

\subsubsection{Guided re-evaluation prompts}
The re-evaluation stage uses chain-of-thought prompting for all guidance conditions.
The only difference across conditions is the guidance modality attached above or injected into the prompt.

\begin{promptbox}{Image-guided CoT re-evaluation prompt}
Context:
<context>

Question: <question>

Your teacher has provided a mind map above to guide your reasoning.

Follow the reasoning process shown in the mind map step by step,
then provide your final answer.

Format your response as:
Reasoning: <follow the mind map step by step>
Answer: <your final answer>
\end{promptbox}

\begin{promptbox}{Text-guided CoT re-evaluation prompt}
Context:
<context>

Question: <question>

Your teacher has provided the following guidance:
<text_guidance>

Follow this reasoning guidance step by step,
then provide your final answer.

Format your response as:
Reasoning: <follow the guidance step by step>
Answer: <your final answer>
\end{promptbox}

\begin{promptbox}{Re-evaluation answer verifier}
You are verifying if a student's answer matches the teacher's validated answer.

Student's Full Response:
<student_raw_output>

Teacher's Validated Answer: <teacher_answer>

Task: Determine if the student's final answer matches the teacher's answer.
- Consider semantic equivalence, not only string match
- If the student clearly states or implies the teacher's answer:
  CORRECT
- Otherwise:
  INCORRECT

End with exactly:
Decision: CORRECT
or
Decision: INCORRECT
\end{promptbox}

\subsubsection{Teacher guidance construction}
The teacher first answers the question in a multi-turn conversation.
Round 2 then rewrites that reasoning into one of four guidance artifacts.
For the image variants, the teacher outputs Graphviz DOT graph code, which is later rendered into an image.
The two image prompts differ only in content style, and the two text prompts mirror the same contrast.

\begin{promptbox}{Teacher image prompt: direct (generate graph code)}
You are an expert teacher creating a reasoning mind map for students.

Based on your reasoning process above,
create a visual mind map that guides students
to reach the correct answer.

CRITICAL:
The mind map must enable students to derive the answer
by following the reasoning path.
- Each node must be easy to understand and actionable
- Break complex reasoning into small digestible nodes

Requirements:
1. Structure:
   - Use < 15 nodes
   - Create a true graph with branches and convergence
     (not a linear chain)
   - Use top-to-bottom layout (rankdir=TB)
2. Node format:
   - Each node label < 10 words
3. Content guidelines:
   - Allowed: specific hints, intermediate conclusions,
     key facts, reasoning steps
   - Forbidden: the final answer itself

Generate only Graphviz DOT code:
digraph ReasoningProcess {
    graph [size="8,8", dpi="150", rankdir=TB];
    node [shape=box, style=rounded];
}
\end{promptbox}

\begin{promptbox}{Teacher image prompt: abstract (generate graph code)}
You are an expert teacher creating a reasoning mind map for students.

Based on your reasoning process above,
create an abstract mind map that teaches students reasoning patterns.

CRITICAL:
The mind map must teach students how to think,
not what to find.
- Each node must be easy to understand and actionable
- Break complex reasoning into small digestible nodes

Requirements:
1. Structure:
   - Use < 15 nodes
   - Create a true graph with branches and convergence
     (not a linear chain)
   - Use top-to-bottom layout (rankdir=TB)
2. Node format:
   - Each node label < 10 words
3. Content guidelines:
   - Allowed: general reasoning strategies,
     meta-questions, logical operations
   - Forbidden: specific hints, intermediate conclusions,
     key facts, the final answer

Generate only Graphviz DOT code:
digraph ReasoningProcess {
    graph [size="8,8", dpi="150", rankdir=TB];
    node [shape=box, style=rounded];
}
\end{promptbox}

\begin{promptbox}{Teacher text prompt: direct}
You are an expert teacher creating reasoning guidance for students.

Based on your reasoning process above,
create text guidance that guides students
to reach the correct answer.

CRITICAL:
The guidance must enable students to derive the answer
by following the reasoning path.
- Each step must be easy to understand and actionable
- Break complex reasoning into small digestible steps

Requirements:
1. Structure:
   - Use < 15 steps
   - Create a reasoning flow with branches and convergence
     (not a linear chain)
2. Content guidelines:
   - Allowed: specific hints, intermediate conclusions,
     key facts, reasoning steps
   - Forbidden: the final answer itself

Generate text guidance.
\end{promptbox}

\begin{promptbox}{Teacher text prompt: abstract}
You are an expert teacher creating reasoning guidance for students.

Based on your reasoning process above,
create abstract text guidance that teaches students reasoning patterns.

CRITICAL:
The guidance must teach students how to think,
not what to find.
- Each step must be easy to understand and actionable
- Break complex reasoning into small digestible steps

Requirements:
1. Structure:
   - Use < 15 steps
   - Create a reasoning flow with branches and convergence
     (not a linear chain)
2. Content guidelines:
   - Allowed: general reasoning strategies,
     meta-questions, logical operations
   - Forbidden: specific hints, intermediate conclusions,
     key facts, the final answer

Generate text guidance.
\end{promptbox}

\subsubsection{Graph-to-text control}
The graph-to-text control is produced by converting the generated DOT graph code into linear text with a separate prompt.
This preserves node content more faithfully than the direct text-guidance prompt, but it removes the visual and spatial structure of the rendered image.

\begin{promptbox}{Graph-to-text conversion prompt}
You are converting a reasoning mind map
(Graphviz DOT code) into a text description.

Requirements:
1. Preserve all nodes.
2. Clearly express relationships with ordered markers.
3. Express branching explicitly.
4. Express convergence explicitly.
5. Do not introduce new content beyond the nodes.
6. Keep the text concise but complete.

Input DOT code:
<graph_code>

Output a text description of the reasoning process.
\end{promptbox}

\subsubsection{Ablation prompts}
The ablation study keeps the same multi-turn teacher conversation and changes only the Round-2 graph-generation prompt.
The cleaner main-text variants are the structure ablation, the 5-node and 10-node node-budget ablations, and the combined 10-node chain ablation.
The direct and abstract versions share the same structural constraints.
Their only difference is whether node content may include concrete hints or must remain abstract.

\begin{promptbox}{Ablation prompt: chain}
Requirements:
1. Structure:
   - Use < 15 nodes
   - Create a strictly linear chain
     (each node connects to exactly one next node)
   - No branching, no parallel paths, no merging
   - Use top-to-bottom layout (rankdir=TB)
2. Node format:
   - Each node label < 10 words
3. Content:
   - Direct version: specific hints and key facts allowed
   - Abstract version: only general reasoning strategies allowed
\end{promptbox}

\begin{promptbox}{Ablation prompt: 5 nodes}
Requirements:
1. Structure:
   - Use exactly 5 nodes or fewer
     (strict limit; do not exceed)
   - Create a true graph with branches and convergence
     (not a linear chain)
   - Use top-to-bottom layout (rankdir=TB)
2. Node format:
   - Each node label < 10 words
3. Content:
   - Direct version: specific hints and key facts allowed
   - Abstract version: only general reasoning strategies allowed
\end{promptbox}

\begin{promptbox}{Ablation prompt: 10 nodes}
Requirements:
1. Structure:
   - Use exactly 10 nodes or fewer
     (strict limit; do not exceed)
   - Create a true graph with branches and convergence
     (not a linear chain)
   - Use top-to-bottom layout (rankdir=TB)
2. Node format:
   - Each node label < 10 words
3. Content:
   - Direct version: specific hints and key facts allowed
   - Abstract version: only general reasoning strategies allowed
\end{promptbox}

\begin{promptbox}{Ablation prompt: 10 nodes + chain}
Requirements:
1. Structure:
   - Use exactly 10 nodes or fewer
     (strict limit; do not exceed)
   - Create a strictly linear chain
     (each node connects to exactly one next node)
   - No branching, no parallel paths, no merging
   - Use top-to-bottom layout (rankdir=TB)
2. Node format:
   - Each node label < 10 words
3. Content:
   - Direct version: specific hints and key facts allowed
   - Abstract version: only general reasoning strategies allowed
\end{promptbox}